\documentclass[a4paper, conference]{IEEEtran}
\IEEEoverridecommandlockouts

\usepackage{cite}
\usepackage{amsmath,amssymb,amsfonts}
\usepackage{algorithmic}
\usepackage{graphicx}
\usepackage{textcomp}
\usepackage{xcolor}
\def\BibTeX{{\rm B\kern-.05em{\sc i\kern-.025em b}\kern-.08em
    T\kern-.1667em\lower.7ex\hbox{E}\kern-.125emX}}

\usepackage{algorithm}

\usepackage[compact]{titlesec}  
\titlespacing{\section}{0pt}{0pt}{0pt}
\titlespacing{\subsection}{0pt}{0pt}{0pt}
\titlespacing{\subsubsection}{0pt}{0pt}{0pt}
\begin{document}

\title{D2D Power Allocation \\via Quantum Graph Neural Network
\thanks{ This work was supported by the IITP(Institute of Information \& Coummunications Technology Planning \& Evaluation)-ITRC(Information Technology Research Center) grant funded by the Korea government(Ministry of Science and ICT)(IITP-2025-RS-2023-00260098) (Corresponding author: Won-Joo Hwang)}
}

\author{
\IEEEauthorblockN{Le Tung Giang}
\IEEEauthorblockA{
\textit{Dept. of Info. Convergence Eng.} \\
\textit{Pusan National University}\\
Busan 46241, Republic of Korea \\
giang.lt2399144@pusan.ac.kr}
\and
\IEEEauthorblockN{Nguyen Xuan Tung}
\IEEEauthorblockA{
\textit{Faculty of Interdisciplinary Digital Technology} \\
\textit{PHENIKAA University}\\
Hanoi 12116, Viet Nam \\
tung.nguyenxuan@phenikaa-uni.edu.vn}
\and
\IEEEauthorblockN{Won-Joo Hwang}
\IEEEauthorblockA{ 
\textit{Center for Artificial Intelligence Research}\\
\textit{Pusan National University}\\
Busan 46241, Republic of Korea \\
wjhwang@pusan.ac.kr}
}

\maketitle

\begin{abstract}
Increasing wireless network complexity demands scalable resource management. Classical GNNs excel at graph learning but incur high computational costs in large-scale settings. We present a fully quantum Graph Neural Network (QGNN) that implements message passing via Parameterized Quantum Circuits (PQCs). Our Quantum Graph Convolutional Layers (QGCLs) encode features into quantum states, process graphs with NISQ-compatible unitaries, and retrieve embeddings through measurement. Applied to D2D power control for SINR maximization, our QGNN matches classical performance with fewer parameters and inherent parallelism. This end-to-end PQC-based GNN marks a step toward quantum-accelerated wireless optimization.

\end{abstract}

\section{Introduction}

Resource management is a key challenge in modern wireless systems due to growing node density and dynamic traffic demands~\cite{2025_TMC_Dan_Semantic, 2023_TMC_Dan_HCFL,tung2024jointly, 2025_Tung_conf}. While classical methods like WMMSE offer near-optimal performance, they scale poorly. Recent works leverage Graph Neural Networks (GNNs) for power allocation by exploiting network structure~\cite{2025_Dan_LARS, 2025_Dan_FedOMG}, using message passing to learn interference patterns~\cite{2025_TVT_Tung_DisGNN, 2024_Sensor_Giang_HGNN, 2023_GNN_Hieu, 2025_TVT_ISAC_Tung}. However, GNNs suffer from memory bottlenecks and over-smoothing on large graphs~\cite{2025_COMST_Tung_GNN}.

Quantum machine learning (QML) offers a promising alternative, leveraging superposition and entanglement to represent complex data efficiently. Parameterized Quantum Circuits (PQCs) allow quantum models to be trained via classical optimizers, often with fewer parameters~\cite{2025_TVT_Hieu_HQCNN}. And recently, Quantum Graph Neural Networks (QGNNs) embed PQCs into GNNs, using quantum parallelism to reduce classical computation~\cite{2025_ICAIIC_Giang_QGNN}.

In this work, we propose a fully quantum GNN architecture for wireless power control in D2D communication. By implementing message passing through PQCs, the model efficiently scales on NISQ devices. Experiments on SINR maximization show that our QGNN achieves competitive performance with improved scalability and quantum efficiency.

\section{System Model and Problem Formulation}

\begin{figure*}[t]
    \centering
    \begin{minipage}[t]{0.5\linewidth}
        \centering
        \includegraphics[width=0.8\linewidth]{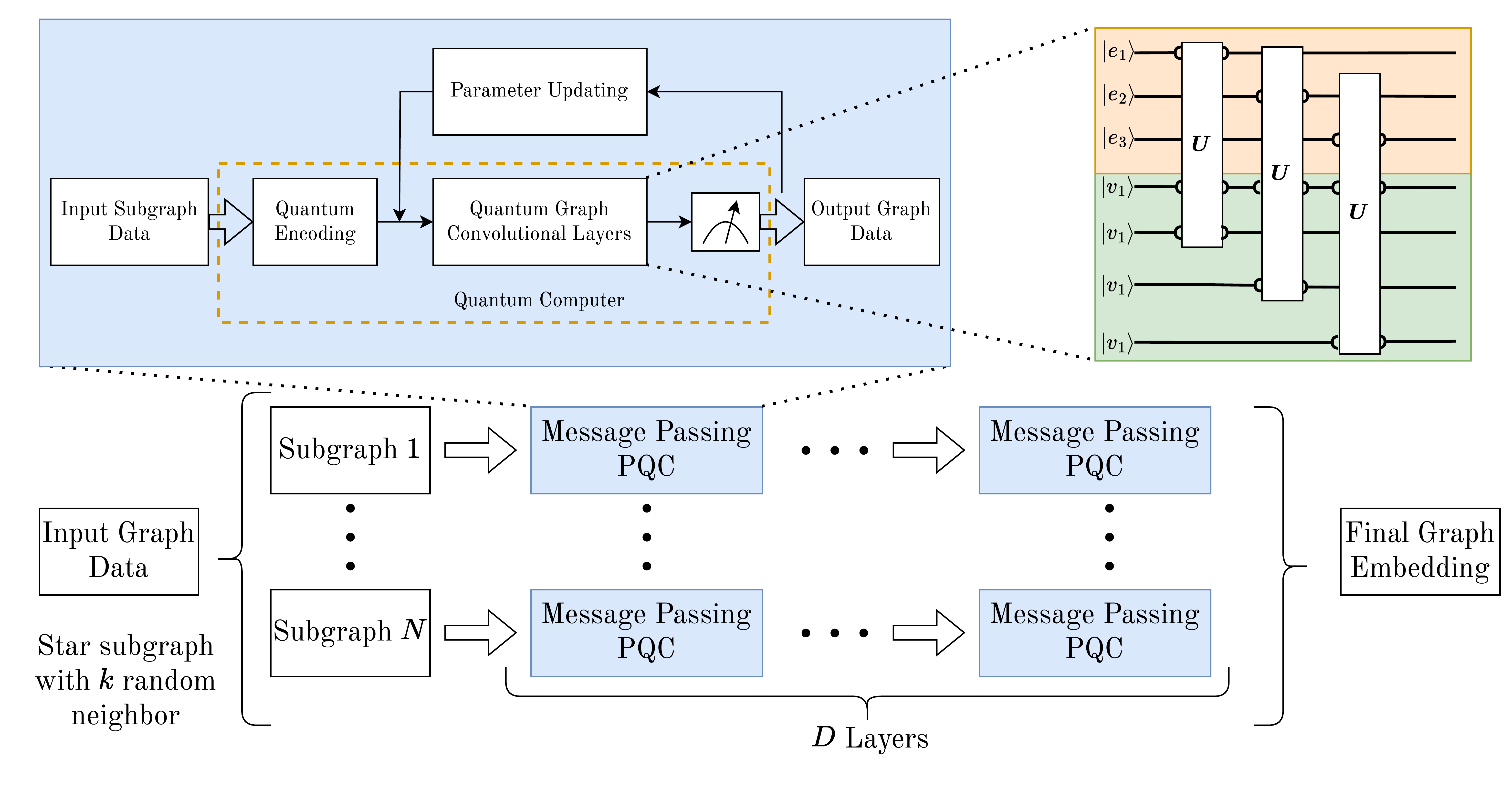}
        \caption{The QGNN Architecture Overview.}
        \label{fig:QGNN_Architecture}
    \end{minipage}
    \hspace{1cm}
    \begin{minipage}[t]{0.4\linewidth}
        \centering
        \includegraphics[width=0.8\linewidth]{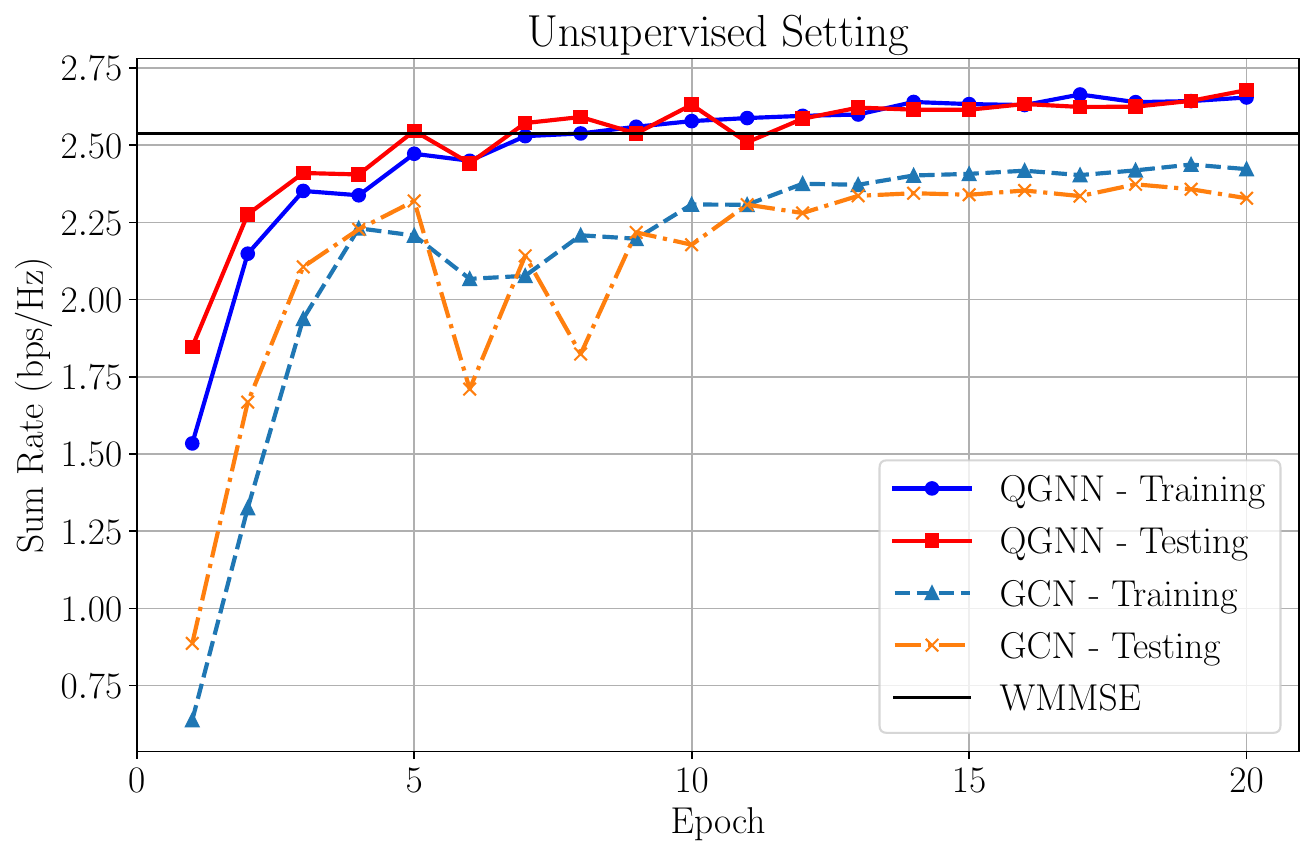}
        \caption{Models performance versus training epochs.}
        \label{fig:results}
    \end{minipage}
\end{figure*}

In this work, we consider a system with $M$ D2D pairs denoted by $\{D_m\}^{M}_{m=1}$  randomly located within a two-dimensional square region of edge length $d$. For each D2D pair $D_m$, we denote its transmitter and receiver as $T_m$ and $R_m$, respectively. Both the transmitter an receiver are equipped with a single antenna, with the distance between each pair lying within the range of  $d_{min}$ and $d_{max}$. Furthermore, all D2D pairs reuse the full bandwidth to transmit and the channel multiplexing is not required in this scenario.
Let $\boldsymbol{p}_k$ denote the power allocated to the $k$-th transmitter. The received signal at receiver $m$ is $y_{m} = g_{mm} p_{m} s_{m} + \sum_{k \neq m}^{M} g_{km} p_{k} s_{k} + n_{m}$,
where $g_{km} \in \mathbb{C}$ denotes the channel state from the $k$-th transmitter to the $m$-th receiver and $n_{m} \sim \mathcal{CN}(0, \sigma_{m}^2)$ denotes the additive noise following the complex Gaussian distribution. Thus, the signel-to-inteference-plus-noise ratio (SINR) for the $m$-th reveicer is expressed as:
\begin{align}
    \gamma_{m} = \frac{\vert g_{mm} p_{m} \vert^2}{\sum_{k\neq m}^{M}\vert g_{km} p_{k} \vert^2 + \sigma_{m}^2}.
\end{align}

Denote $\boldsymbol{p} = [p_1,\ldots,p_m]^T \in \mathbb{C}^{K}$ as the power allocation vector. The objective is to maximize the weighted total SINR by finding the optimal power allocation vector, which can be formulated as
\begin{equation}
\label{eq: w_sum}
\begin{aligned}
\underset{\boldsymbol{p}}{\textrm{maximize}} \quad & \sum_{m=1}^{M} \alpha_{m}\log_2(1 + \gamma_{m}) \\
    \textrm{s.t.} \:\qquad & 0\leq p_{m} \leq P_{
    \textrm{max}}, \forall m,
\end{aligned}
\end{equation}
where $\alpha_{m}$ is the weight for the $m$-th transceiver pair.

\section{Methodology}

We propose a hybrid quantum-classical framework where message passing is performed by parameterized quantum circuits using unitary operations instead of weight matrices.
As shown in Fig.~\ref{fig:QGNN_Architecture}, the graph is decomposed into star subgraphs, each processed by a Quantum Graph Convolutional Layer (QGCL) that updates the central node based on its neighbors.


\subsection{Graph Decomposition}

To handle the qubit limitations of current quantum devices, we propose a stochastic graph decomposition based on $k$-neighbor star subgraphs. Leveraging the random neighbor sampling, and the local nature of GNN updates, we decompose a graph $\mathcal{G}$ with $N$ nodes into $N$ overlapping stars. Each node serves as the center of one star, with $k$ neighbors randomly selected as leaves. This ensures full coverage of the graph, with each node acting as a center once and possibly as a leaf in others, preserving local structure for quantum message passing.

\subsection{Message Passing Implementing PQC}

Motivated by the message passing mechanism in classical GNNs, we design a Parameterized Quantum Circuit (PQC) that directly implements a similar operation using basic quantum gates.
For the design of the quantum graph convolution layer (QGCL), we propose the following process:
For each star subgraph, we apply a shared unitary operator to process the center node, a leaf node, and their connecting edge,
each node aggregates incoming messages via a quantum-compatible operation (e.g., unitary over concatenated states).
After $L$ layers, final node embeddings can be used for downstream tasks. For graph-level tasks, we apply pooling function such as sum or mean.
Fig.~\ref{fig:QGNN_Architecture} illustrates an example with $4$ nodes and $6$ edges, each node represented by a 2D feature vector and encoded into a $7$-qubit PQC.

\section{Numerical Results}
In this section, we evaluate the proposed QGNN on the D2D power‐allocation task. We adopt the GCN architecture from \cite{2021_JSAC_Shen_GNN} as our classical benchmark. Both models are trained in an unsupervised fashion using only $300$ channel realizations. Training is performed with the Adam optimizer and an initial learning rate of $\eta = 5\times10^{-2}\,$.
Fig.~\ref{fig:results} illustrates the performance of both models versus the training epochs, compared to the WMMSE power allocation. As shown, the QGNN rapidly surpasses the WMMSE baseline by epoch~$4$ and continues improving to achieve approximately $2.67\,$bps/Hz-about a $6\%$ gain over the GCN, which plateaus at the WMMSE level. Moreover, the QGNN’s training and testing curves align closely, indicating both faster convergence and superior generalization compared to the classical GCN.

\section{Conclusion}

We propose a fully quantum Graph Neural Network (QGNN) for D2D power allocation, implementing message passing end-to-end via parameterized quantum circuits. In an unsupervised SINR-maximization task on just 300 channel realizations, QGNN outperforms both the classical GCN benchmark and WMMSE-surpassing WMMSE at very early stage of epoch~$4$, converging faster, and achieving a final sum-rate gain of~$\approx6\%$-while exhibiting tight train/test alignment and strong generalization. This proof-of-concept demonstrates the promise of quantum-native graph learning for wireless resource management; future work will address larger topologies, quantum error mitigation on NISQ hardware, and hybrid training schemes.

\bibliographystyle{IEEEtran}
\bibliography{reference}

\end{document}